\documentclass[12pt]{article}
\usepackage{amsmath,amssymb}
\usepackage{graphicx}
\usepackage{enumerate}
\usepackage[sectionbib]{natbib}
\usepackage{url} 

\newcommand{\blind}{1}

\renewcommand{\clearpage}{}
\usepackage{fullpage}
\usepackage{upgreek}
\usepackage{dsfont}
\usepackage{float}
\usepackage[normalem]{ulem}
\usepackage{hyperref}
\usepackage{xcolor,soul,tcolorbox}
\usepackage{bbm}
\usepackage{amsthm}
\usepackage{enumitem}
\newtheorem{theorem}{Theorem}

\newtheorem{corollary}{Corollary}
\newtheorem{assumption}{Assumption}

\theoremstyle{remark}

\usepackage{verbatim}
\usepackage{xspace}
\usepackage{multirow}
\usepackage[flushleft]{threeparttable}
\usepackage{algorithm}
\usepackage{algpseudocode}

\def\S{\mathcal{S}}

\def\EE{\mathbb{E}}

\def\G{\boldsymbol{G}}
\def\g{g^*_\pi}
\def\gn{\hat g_{n, \pi}}
\def\F{\mathcal{F}}
\def\T{\mathcal{T}}

\def\goes{\rightarrow}

\def\A{\mathcal{A}}
\def\D{\mathcal{D}}

\def\R{\mathbb{R}}
\def\Pn{\mathbb{P}_n}

\def\P{\mathcal{P}}

\def\leqconst{\lesssim}

\newcommand{\norm}[1]{\|#1\|}
\newcommand{\bignorm}[1]{\big \|#1 \big \|}

\newcommand{\abs}[1]{|#1|}

\newcommand{\bigdkh}[1]{\big\{#1 \big\}}

\newcommand{\indicator}[1]{ \mathds{1}_{\{#1\}}}

\def\EE{\mathbb{E}}

\def\R{\mathbb{R}}

\def\ones{\mathbf{1}}

\def\argminb{\operatorname*{argmin}}

\def\wcvg{\Rightarrow}

\def\Q{\mathcal{Q}}
\def\given{\, | \,}
\def\Given{\, \Big| \,}

\def\transpose{\top}

\renewcommand{\G}{\mathcal{G}}

\begin{document}

\def\spacingset#1{\renewcommand{\baselinestretch}%
	{#1}\small\normalsize} \spacingset{1}


\if1\blind
{
	\title{\bf Off-Policy Estimation of Long-Term Average Outcomes with Applications to Mobile Health}
	\author{Peng Liao\hspace{.2cm}\\
		Department of Statistics, University of Michigan\\ \\
		Predrag Klasnja \\
		School of Information, University of Michigan\\ \\
		Susan Murphy \\
		Department of Statistics, Harvard University}
	\date{}
	\maketitle
	
} \fi

\if0\blind
{
	\bigskip
	\bigskip
	\bigskip
	\begin{center}
		{\LARGE\bf Off-Policy Estimation of Long-Term Average Outcomes with Applications to Mobile Health}
	\end{center}
	\medskip
} \fi

\bigskip
\begin{abstract}
	
	Due to the recent advancements in wearables and sensing technology, health scientists are increasingly developing mobile health (mHealth) interventions. In mHealth interventions, mobile devices are used to deliver treatment to individuals as they go about their daily lives.  These treatments are generally designed to impact a near time, proximal outcome such as stress or physical activity. The mHealth intervention policies, often called just-in-time adaptive interventions, are decision rules that map a individual's current state (e.g., individual's past behaviors as well as current observations of time, location, social activity, stress and urges to smoke) to a particular treatment at each of many time points. The vast majority of current mHealth interventions deploy expert-derived policies. In this paper, we provide an approach for conducting inference about the performance of one or more such policies using historical data collected under a possibly different policy. Our measure of performance is the average of proximal outcomes over a long time period should the particular mHealth policy be followed. We provide an estimator as well as confidence intervals. This work is motivated by HeartSteps, an mHealth physical activity intervention.

\end{abstract}

\noindent%
\vfill

\newpage
\spacingset{1.5} 

\begingroup
\allowdisplaybreaks
\section{Introduction}

Due to the recent advancement in mobile device and sensing technology, health scientists are more and more  interested in developing mobile health (mHealth) interventions. In mHealth,  mobile devices  (e.g., wearables and smartphones) are used to deliver interventions to individuals as they go about their daily lives. In general, there are two types of mHealth treatments. Most are “pull” treatments that reside on the individual's mobile device and allow the individual to access treatment content as needed. This work  focuses on the second type, the ``push''  treatment, typically in the form of a notification or a text message that appears on a mobile device.  There is a wide variety of possible treatment messages  (e.g., behavioral, cognitive, and motivational message and reminders). These treatments are generally intended to impact a near time, proximal outcome, such as stress or behaviors such as physical activity over some subsequent minutes/hours.    The mHealth intervention policies, often called just-in-time adaptive interventions in the mHealth literature \citep{nahum2018just}, are decision rules  that map the individual’s current state (e.g., past behaviors as well as current observations of location, social activity, stress and urges to smoke) to a particular treatment at each of many time points.  Many mHealth interventions are designed for long-term use in chronic disease management \citep{lee2018effective}. The vast majority of current mHealth interventions deploy expert-derived policies with limited use of data evidence (for an example see \cite{kizakevich2014personal}), however the long-term efficacy of these policies on the health behavior is not well understood. An important first step toward developing data-based, effective mHealth interventions is to properly measure the long-term performance of these policies. In this work, we provide an approach for conducting inference about the optimality of one or more mHealth policies of interest. Our optimality criterion is the long-term average of the proximal outcomes should a particular mHealth policy be followed. We develop a flexible method to estimate the performance of an mHealth policy using a historical dataset in which the treatments are decided by a possibly different policy.

This work is motivated by HeartSteps \citep{klasnja2015microrandomized}, an mHealth physical activity intervention. To design this intervention, we are conducting a series of studies.  The first, already completed, study was for 42 days.  The last study will be for one year.   Here we focus on the intervention component involving activity suggestions. These suggestions may be delivered at each of the five individual-specified times per day. While in the first study there were $42\times5=210$ time points per individual, in the year-long study there will be about 2,000 time points per individual. The proximal outcome is the step count in the 30 minutes following each of the five times per day.    Our goal is to use the data collected from the first 42-day study to predict and estimate the long-term average of proximal outcomes for a variety of policies that could be used to decide whether or not to send the  activity suggestion at each time point in the year-long study.  The  42-day study was a Micro-Randomized Trial (MRT) \citep{klasnja2015microrandomized,Liaoetal2015}.   In an MRT, a known stochastic policy, also called a behavior policy, is  used to decide when and which type of treatment to provide at each time point.  A partial list of MRTs in the field or completed can be found at the website\footnote{\url{http://people.seas.harvard.edu/~samurphy/JITAI_MRT/mrts4.html}}.  From an experimental point of view, the stochastic behavior policy is used to conduct sequential randomizations within each individual.  Here the adjective, ``stochastic'', means that at each time point each individual is randomized between the possible treatments.  In this work we focus on settings in which the randomization probabilities are  known functions of the individual's past data; this is the case with MRTs by design. 

The rest of the article is organized as follows. Section \ref{section: mdp} provides  a review of Markov Decision Processes. In Section \ref{section:related work}, we review related work. Section \ref{section: off-policy evaluation} develops an estimator for the long-term average proximal outcome; then  Section \ref{section: theory} provides  the asymptotic distribution of this estimator. As the estimation requires tuning parameters, in Section \ref{section: simulation} we provide a procedure to select the tuning parameters.  Simulations are used to assess the coverage probability of the proposed confidence intervals in various settings.  A case study using data from the 42-day MRT of HeartSteps is presented in Section \ref{section: heartsptes}.  We end with a discussion of future work in Section \ref{sec: discussion}.

\section{Distributional Assumptions and Goal}

\label{section: mdp}

The data for each individual is of the form
$$\D= \{S_1, A_1, S_2, A_2, S_3, \dots, S_t, A_t, S_{t+1}, \dots, S_{T+1}\},$$
where $t$ indexes  time points, $S_{t} \in \S$ is the individual's state and $A_t \in \A$ is the treatment (usually called the action) assigned at time, $t$.   The action space, $\A$, is discrete and finite.  In mHealth, the state, $S_{t}$, contains the
time-varying information (e.g., current location) as well as summaries
of historical data up to and including time, $t$ (e.g., summaries of previous physical activity).  The actions are different types of treatments that are delivered to the individual via a smartdevice; these treatments can be reminders, motivational messages, messages prompting self-reflection and so on.  For simplicity, we assume that the duration over which data is collected, $T$, is non-random and same for all individuals.  	The proximal outcome (also called the reward), denoted by $R_{t+1} \in \R$, is assumed to be a known function of  $(S_{t}, A_t, S_{t+1})$.  In mHealth, the reward is often chosen to measure the near-term impact of the current action (e.g., the number of steps in a pre-specified time window  after each time point).   In this work, we focus on the case of continuous rewards (see Section \ref{sec: discussion} for a discussion about other types of rewards).   In HeartSteps, the binary action is whether  an activity suggestion is delivered and the reward is the 30-min step count following each decision time.

We assume that the distribution of the states satisfies the Markovian property, that is,  for $t \geq 1$, $S_{t+1} \perp \{S_1, A_1, \dots, S_{t-1}, A_{t-1}\} \given \{S_t, A_t\}$. Furthermore, we assume that the conditional distribution (also called the transition kernel) of  $S_{t+1}  \given \{S_t, A_t\}$ is time-homogeneous.   Denote the transition kernel by $P$; thus given a measurable set, $B$, in the state space, $\S$,  $P(B \given s, a) = \Pr(S_{t+1}\in B\given S_t=s, A_t=a)$.  Note that $P$ does not depend on $t$ due to the above time-homogeneity assumption.  Denote by $p(s'\given s, a)$ the transition density with respect to some reference measure on $\S$ (e.g., counting measure when $\S$ is discrete).  	Let $r(s, a)$  denote the conditional expectation of the reward given state and action, i.e., $r(s, a) = \EE(R_{t+1}\given S_t = s, A_t = a)$.  	The tuple, $(\S,\A, P)$, is called a Markov Decision Process (MDP) \citep{howard1960dynamic,puterman1994markov,sutton2018reinforcement}. 
In mHealth, non-stationarity in $P$ is likely to occur if there are unobserved aspects of the current state (e.g., individual's engagement and/or burden). Therefore, practically, it is critical to strive to collect sufficient information  (via self-report or wearable sensors) to  represent individual's state. 

Note that the MDP does not specify the distribution of the actions.   And indeed the distribution of the actions may not satisfy the Markovian property.  In an MRT, the actions, $\{A_t\}_{t=1}^T$, are randomized with probabilities that can depend on the entire history  prior to time point $t$, $H_t = \{S_1, A_1, \dots, S_t\}$.  Denote the distribution of $A_{t} \given H_t$ by $\pi_{t}^b(\cdot \given H_t)$.  We call $\pi_b=\{\pi_{1}^b,\ldots,\pi_{T}^b\}$, a stochastic behavior policy. Throughout we assume that $\pi_b$ is known (as is the case in an MRT) and  that the  probabilities are strictly positive, i.e., $\pi_{t}^b(a\given H_t) \geq p_{\min} > 0$ for all $a \in \A$, $H_t$ and  $t\le T$.

Suppose that a pre-specified time-invariant, Markovian policy, $\pi$, is being considered for use in future.  Our goal is to conduct inference for  the resulting average of the rewards over a large number of time points.  In mHealth, the policy  might be an expert-constructed policy.   Considering a long time period makes most sense    for individuals who are struggling with chronic problems or disorders for which, at this time, there is no general cure.  Many health-behavior problems fall into this area including obesity, hypertension, adherence to medications for AIDs, mental illness and addictions.  
Let $\pi(a\given s)$ be the probability of choosing the action, $a$, at the state, $s$.  Given a dataset that consists of $n$ independent, identically distributed (i.i.d.)~observations of $\D$, we aim to estimate the average reward of the policy, defined as
\begin{align}
\eta^\pi (s) = \operatorname*{limsup}_{t^*\goes \infty} \EE_{\pi} \left(\frac{1}{t^*} \sum_{t=1}^{t^*} R_{t+1} \Given S_1= s\right),
\label{def: avg rwrd}
\end{align}
where the expectation, $\EE_{\pi}$, is taken over the trajectory $\{S_1, A_1, S_2, \dots, S_{t^*}, A_{t^*}, S_{t^*+1}\}$ in which the actions are selected according to the policy, $\pi$, that is, the likelihood in the expectation is given by $\indicator{S_1 = s}\prod_{t=1}^{t^*}\pi(A_t\given S_t) p(S_{t+1}\given S_{t}, A_t)$.  The policy, $\pi$, induces a Markov chain on the state with the transition kernel, $P^\pi(\cdot\given s) = \sum_{a \in \A} \pi(a\given s) P(\cdot\given s, a)$.

Suppose for now the state space, $\S$, is finite. It is known that the limit in (\ref{def: avg rwrd}) exists, i.e., $\eta^\pi (s) = \operatorname*{lim}_{t^*\goes \infty} \EE_{\pi} (\frac{1}{t^*} \sum_{t=1}^{t^*} R_{t+1} \given S_1= s)$ (Theorem A.6 on p.~595 in \cite{puterman1994markov}). Furthermore, when the induced Markov chain, $P^\pi$, is irreducible, the average reward is independent of initial state and is given by
\begin{align}
\eta^\pi(s) = \eta^\pi = \sum_{s, a}\pi(a\given s) d^\pi(s) r(s, a), \label{constant avg reward}
\end{align}
where $d^\pi(\cdot)$ is the stationary distribution. The existence of stationary distribution is guaranteed by the irreducibility assumption on $P^\pi$ (\cite{puterman1994markov}, p.~592).  The above results can be extended to general state spaces (e.g., $\S \subset \R^d$) under more involved conditions on the transition kernel, $P^\pi$, analogous to the finite state case (see, for example,  \cite{hernandez2012further}, chap.~7).~Practically the irreducibility  assumption implies that time-invariant information cannot be  included in the state. Motivated by  mHealth applications, in Supplement A we present a generalization that allows the average reward to depend on time-invariant variables.  In the case of mHealth, time-invariant variables might be gender, baseline severity, genetics and so on.

We propose to conduct inference about the long-term performance of each policy, $\pi$, via its average reward, $\eta^\pi$.   This is because the average reward, $\eta^\pi$,  is an asymptotic surrogate of the average of finite rewards over a long period of time. In fact, it can be shown that $$\sup_{s \in S} \left\{\Big|\EE_\pi \left(\frac{1}{t^*}\sum_{t=1}^{t^*} R_{t+1}\Given S_1 = s\right) - \eta^\pi\Big|\right\} = O(1/t^*),$$ where the leading constant depends on the mixing time of $P^\pi$ (see Theorem 7.5.10 in \cite{hernandez2012further}).~In the case of HeartSteps, the goal is to use the data from the 42-day MRT study to estimate the average reward, $\eta^\pi$, for a variety of policies $\pi$.   The average reward, $\eta^\pi$, provides a  proxy for the average of the 30-min step counts when the policy, $\pi$, is used to determine whether to send the activity suggestions over a long time period (e.g., a year: $5\times 365$ time points).

Note that the data, $\cal D$, on each individual includes observations over $T$ time points  and the actions are selected according to a behavior policy.  
However, as will be seen, the above assumptions including the Markovian and irreducibility assumptions will allow us to estimate the average reward over a long time period and under a different policy.

Lastly as mentioned above the focus here is to conduct inference for the long run average of equally weighted rewards under the target policy using data collected under a possibly different policy.    An alternate, and more common, inference target is based on an expected discounted sum of rewards, $ \EE_{\pi} (\sum_{t=1}^{\infty} \gamma^{t-1} R_{t+1} \given S_1 = s)$, with the discount rate, $\gamma\in[0,1)$.   When the discount rate, $\gamma$, is small (e.g., $\gamma = 0.5$), the discounted sum of rewards focuses only on finitely many near-term rewards.   Note that even with a large discount rate of $\gamma=0.99$, the reward at $t=100$ has a weight of $0.37$ and the reward $t=200$ has a weight of $0.13$. Recall our motivating mHealth intervention is being designed to optimize the overall physical activity over one year.  From a scientific point of view, the rewards in the distant future are as important as the near-term ones, especially when considering the effect of habituation and burden.   With this in mind, we opt for the long-term average reward, which can be viewed as a proxy for the (undiscounted) average of rewards over a long period of time.   In fact, the conditional expectation of the sum of discounted rewards is related to the average reward; as $\gamma \goes 1$, the above conditional expectation of the sum of the discounted rewards normalized by the constant, $1/(1-\gamma)$, converges to the average reward, $\eta_\pi$ \citep{mahadevan1996average}. In the online setting many researchers focus on a discounted sum of rewards. This is because the Bellman operator, for the expected  discounted sum of rewards,  is a contraction \citep{sutton2018reinforcement}; the contraction provides greater  computational stability and simpler convergence arguments.   However, as we shall see below, consideration of the average reward is not problematic in the batch  (i.e., off-line) setting. 

\section{Related Work}
\label{section:related work}

The evaluation of a given target policy using data collected from a different policy (i.e., the behavior policy) is called off-policy evaluation. This has been widely studied in both the statistical  and reinforcement learning (RL) literature.  Many authors have evaluated and contrasted policies in terms of the expected sum of rewards over a  finite number of time points \citep{murphy2001marginal,chakraborty2013statistical,jiang2015doubly}. However, because these methods often use products of weights with probabilities from the behavior policy in the denominator, the extension to problems with a large number of time points often suffers from a large variance \citep{thomas2016data,jiang2015doubly}.

The most common off-policy evaluation methods for infinite-horizon problems (i.e., a large number of time points) focus on a discounted sum of rewards and are thus based in some way on the value function (in the discounted reward setting $ \EE_{\pi} ( \sum_{t=1}^{\infty} \gamma^{t-1} R_{t+1} \given S_1 = s)$ considered as a function of $s$ is the value function).~\cite{farahmand2016regularized} proposed a regularized version of Least Square Temporal Difference~\citep{bradtke_linear_1996} and statistical properties  were studied. They used a non-parametric model to estimate the value function and derived the convergence rate when training data consists of i.i.d.~transition samples in the form of  state, action, reward and next state.  From a technical point of view, our estimation method is similar to \cite{farahmand2016regularized}, albeit focused on the average reward; most importantly our method relaxes the assumption that Bellman operator can be modeled correctly for each candidate relative value function and only assumes the data consists of i.i.d.~samples of trajectories.    \cite{luckett2019estimating} also focused on the discounted reward setting.  They evaluated policy, $\pi$, based on an average  of $ \EE_{\pi} (\sum_{t=1}^{\infty} \gamma^{t-1} R_{t+1} \given S_1 = s)$ with respect to a pre-selected reference distribution of the state.   While the reference distribution can be naturally chosen as the distribution of the initial state \citep{pmlr-v80-farajtabar18a,liu2018breaking,luckett2019estimating,thomas2016data}, choosing a ``right'' discount rate, $ \gamma$,  can be non-trivial, at least in mHealth.  They assumed a parametric model for  the value function and developed a regularized estimating equation.  In computer science literature, there also exists many off-policy evaluation methods for the discounted reward setting. We refer the interested reader to the recent works by \cite{pmlr-v80-farajtabar18a} and \cite{kallus2019intrinsically} and references therein.

Closest to the setting of this work is the recent work by  \cite{murphy2016batch} and \cite{liu2018breaking}.   \cite{murphy2016batch} considered the average reward setting. They assumed a linear model for the value function and constructed the estimating equations to estimate the average reward.  However the linearity assumption of the value function is unlikely to hold in practice and difficult to validate (e.g., the value function involves the infinite sum of the  rewards). Our method allows the use of a non-parametric model for the value function to increase  robustness.  \cite{liu2018breaking} also considered the average reward and proposed an estimator for the average reward based on estimating the ratio of the stationary distribution under the target policy divided by the stationary distribution under the behavior policy.  However they did not provide confidence intervals or other inferential methods besides an estimator for the average reward.    In addition, they restricted the behavior policy to be Markovian and time-stationary.  In mHealth, the behavior policy can be determined by an algorithm based on the accruing data and thus violates this assumption \citep{liao2018just,dempsey2017stratified}.

\section{Estimator for Off-Policy Evaluation}
\label{section: off-policy evaluation}

We assume that the dataset, $\D_n$, consists of $n$ trajectories: $$\D_n = \left\{\D^i\right\}_{i=1}^n = \left\{S_1^i, A_1^i, S_2^i, \dots, S_T^i, A_T^i, S_{T+1}^i\right\}_{i=1}^n.$$  Each trajectory, $\D^i$, is an i.i.d.~copy of $\D$ described in  Section \ref{section: mdp}.  Recall that $\left\{A_t\right\}_{t=1}^T$, the actions in $\D$, are selected by the behavior policy, $\pi_b$. In the following, the expectation, $\EE$, without the subscript is with respect to the distribution of the trajectory, $\D$, under the behavior policy.  

Below we introduce the estimator for $\eta^\pi$.  We follow the so-called ``model-free'' approach (i.e., does not require modeling the transition kernel, $P$) to estimate the average reward. Our estimator is based on the Bellman equation, also known as the Poisson equation \citep{puterman1994markov}; as will be discussed below this equation characterizes the average reward. 

First consider the setting where the state space, $\S$, is finite and the induced Markov chain, $P^\pi$, is irreducible. Recall that in this setting the average reward, $\eta^\pi$, is a constant given in (\ref{constant avg reward}). Define the relative value function by
\begin{align}
Q^\pi(s, a) =  \operatorname*{lim}_{t^* \goes \infty} \frac{1}{t^*}\sum_{t=1}^{t^*} \EE_\pi\left[\sum_{k=1}^{t} \left\{R_{k+1} - \eta^\pi\right\} \Given S_1 = s, A_1 = a\right]; 
\label{value function. simplified}
\end{align}
this limit is well-defined (\cite{puterman1994markov},  p.~338). If the induced Markov chain is aperiodic, then the relative value function (\ref{value function. simplified}) can be expressed as  $Q^\pi(s, a)   = \EE_\pi\{\sum_{t=1}^{\infty} (R_{t+1} - \eta^\pi) \given S_1 = s, A_1 = a\}$.  The relative value function, $Q^\pi(s, a)$, measures the difference between the expected cumulative rewards under policy $\pi$ and the average reward when the initial state is $s$ and the first action is $a$.  It is easy to verify from the definition that $(\eta^\pi, Q^\pi)$ is a solution of the Bellman  equation:
\begin{align}
\EE_\pi \left\{R_{t+1} + Q(S_{t+1}, A_{t+1}) \given S_t = s, A_t = a\right\}= \eta + Q(s, a),~ \forall (s, a) \in \S \times \A. \label{Bellman equationL eta,V}
\end{align} 
Furthermore, when the induced Markov chain is irreducible, the Bellman equation (\ref{Bellman equationL eta,V}) uniquely identifies the average reward, $\eta^\pi$, and identifies the relative value function, $Q^\pi$,  up to a constant (see \cite{puterman1994markov}, p.~343 for details). That is, the set of the solutions of the Bellman equation (\ref{Bellman equationL eta,V}) is given by $\{(\eta^\pi, Q): Q = Q^\pi + c \ones, c \in \R, \ones(s, a) = 1 \}$.  These results can be generalized to general state spaces (see chap.~7 in \cite{hernandez2012further}). The key assumption for the method proposed here  is  as follows.

\begin{assumption} \label{assumption: irreducible}
	The average reward of the target policy, $\pi$, is independent of state and satisfies (\ref{constant avg reward}).  $(\eta^\pi, Q^\pi)$ is the unique solution of the Bellman equation (\ref{Bellman equationL eta,V}) up to a constant for $Q^\pi$.  The stationary distribution of the induced transition kernel, $P^\pi$, exists. 
\end{assumption}

As the focus of this work is to estimate the average reward,  it will be sufficient to estimate a specific version of $Q^\pi$. Define the shifted relative value function by $\tilde Q^\pi(s, a) = Q^\pi(s, a) - Q^\pi(s^*, a^*)$ for  a specific state-action pair, $(s^*, a^*)$. Obviously  $\tilde{Q}^\pi(s^*, a^*) = 0$ and $\tilde{Q}^\pi(s_1, a_1) - \tilde{Q}^\pi(s_2, a_2) = Q^\pi(s_1, a_1) - Q^\pi(s_2, a_2)$, that is,  the difference in the relative value remains the same.~By restricting the relative value function to satisfy $Q(s^*, a^*) =0$,  the solution of Bellman equation (\ref{Bellman equationL eta,V}) is unique and given by $(\eta^\pi, \tilde Q^\pi)$. 

In the following, we  assume that $\tilde Q^\pi \in \Q$, where $\Q$  denotes a vector space of functions on the state-action space $\S\times \A$ such that $Q(s^*, a^*) = 0$ for all $Q \in  \Q$.  The Bellman operator, $\T_\pi$, with respect to the target policy $\pi$ is given by 
\begin{align}
\T_\pi(s,a; Q) = 
\EE\left\{R_{t+1} + \sum_{a'} \pi(a' \given S_{t+1}) Q(S_{t+1}, a') \Given S_t = s, A_t = a\right\}.
\label{BelErr Op}
\end{align}

Note that  the above conditional expectation does not depend on the behavior policy due to the conditioning on  current state and action. The Bellman error at $(s,a)$ with respect to $(\eta, Q)$ and $\pi$ is defined as $\T_\pi(s,a; Q)-\eta-Q(s,a)$.  From the Bellman equation, this error is zero for all $(s, a)$ when $\eta=\eta^\pi $ and $Q= \tilde Q^\pi$.

Note that the Bellman operator (\ref{BelErr Op}) involves the (unknown) transition kernel, $P$.
Suppose for now that $P$ is known and thus the Bellman operator is known. Since the Bellman error is zero at $\eta=\eta^\pi $ and $Q= \tilde Q^\pi$, a natural way to estimate $(\eta^\pi, \tilde Q^\pi)$ is to minimize the empirical squared Bellman error, i.e.,
\begin{align}
\min_{(\eta, Q) \in \R \times \Q} \Pn \left[\frac{1}{T}\sum_{t=1}^T \left\{\T_\pi(S_t,A_t; Q)-\eta-Q(S_t,A_t)\right\}^2\right], \label{infeas estimator}
\end{align}
where  $\Pn f(\D) = (1/n) \sum_{i=1}^n f(\D^i)$ is the empirical mean over the training data, $\D_n$, for a function of the trajectory, $f(\D)$.  Obviously, this is not a feasible estimator as we don't know the transition kernel and thus  $\T_\pi(S_t,A_t; Q)$ is unknown.   A natural idea is to replace the Bellman operator by its sample counterpart, i.e., replace $\T_\pi(S_t, A_t; Q)$ by $R_{t+1} + \sum_{a'} \pi(a'\given S_{t+1}) Q(S_{t+1}, a')$ in the objective function of (\ref{infeas estimator}).  Unlike the regression problem in which the dependent variable is fully observed, the dependent variable here is $R_{t+1} +  \sum_{a'} \pi(a'\given S_{t+1}) Q(S_{t+1}, a')$, which involves the unknown relative value function, $Q$. As a result, this natural plug-in estimator is biased (see \cite{antos_learning_2008} for a similar discussion in the discounted reward setting).

The above argument motivates a coupled estimator in which we use the estimated Bellman error to form an objective function. In particular,  for each $(\eta, Q)$, we replace the Bellman error,  $\T_\pi(S_t,A_t; Q)-\eta-Q(S_t,A_t)$,  in (\ref{infeas estimator}) by an estimate of the ``projection'' of the Bellman error into a second  function class, $\G$:
\begin{align}
\g (\cdot, \cdot; \eta, Q)=  \argminb_{g \in \G} \EE \left[\frac{1}{T} \sum_{t=1}^{T} \left\{\T_\pi(S_t, A_t;  Q)-\eta- Q(S_t,A_t) - g(S_t, A_t)\right\}^2 \right]. \label{surrogate BelErr Op}
\end{align} 
Throughout we assume the solution of the above optimization exists and is in $\G$ and we call $\g(\cdot, \cdot; \eta, Q)$ a projection for simplicity.  Recall that members of  $\Q$  satisfy $Q(s^*, a^*) = 0$.  A similar constraint needs not be placed on the members of $\G$. It is worth noting that we do not require the assumption that the Bellman error is modeled correctly by $\G$, that is, $\T_\pi(\cdot, \cdot; Q)-\eta-Q(\cdot,\cdot)$ may not be in $\G$.   A natural choice of $\G$ is $\R \oplus \Q = \{ c+ Q: c  \in \R, Q \in \Q \}$, however this is not mandatory.  \cite{farahmand2016regularized} assumed that the Bellman error (in discounted setting) is in fact in $\Q$ in order to develop a non-parametric estimator for the value function.   As we will see, the assumption that the Bellman error belongs to $\G$ is in fact not necessary and can be relaxed (our proof will use the weaker assumption \ref{assumption: G requirement} in Section \ref{section: theory}). The key reason why the projected Bellman error (\ref{surrogate BelErr Op}) allows us to identify $(\eta^\pi, \tilde Q^\pi)$ is because $g_\pi^*(\cdot, \cdot; \eta^\pi, \tilde Q^\pi) = 0$ (see also \textit{(iii)} in Assumption~\ref{assumption: G requirement} ).

We now formally introduce the estimator for $(\eta^\pi, \tilde Q^\pi)$.  This
estimator is designed to minimize the projected Bellman error (\ref{surrogate BelErr Op}). Specifically, the estimator, $(\hat \eta_n^\pi, \hat Q_n^\pi)$, of  $(\eta^\pi, \tilde Q^\pi)$,  is found by solving a coupled (or nested) optimization problem: 
\begin{align}
\min_{(\eta, Q) \in \R \times \Q} \Pn \left\{\frac{1}{T} \sum_{t=1}^{T} \gn^2(S_t, A_t; \eta, Q)\right\} + \lambda_n J_1^2(Q), \label{estimator}
\end{align}
where for each $(\eta, Q)$, $\gn(\cdot, \cdot; \eta, Q)$ is an estimator for the  projection of the Bellman error given by
\begin{align}
\gn(\cdot, \cdot; \eta, Q) =  \argminb_{g \in \G}  \Pn \Bigg[ \frac{1}{T}\sum_{t=1}^{T}  \big \{ R_{t+1} + & \sum_{a'} \pi(a'\given S_{t+1}) Q(S_{t+1}, a') \notag \\
& -\eta- Q(S_t,A_t) - g(S_t, A_t) \big \} ^2  \Bigg ] + \mu_n J_2^2(g). \label{est Bellman error}
\end{align}
where $J_1: \Q \goes \R^+$ and $J_2: \G \goes \R^+$ are two regularizers and $\lambda_n$ and $\mu_n$ are tuning parameters.  

We can see that for every $(\eta, Q)$, $\gn(\cdot, \cdot; \eta, Q)$ is a penalized estimator for the projected Bellman error $\g(\cdot, \cdot; \eta, Q)$ in (\ref{surrogate BelErr Op}).  On the other hand, the objective function in (\ref{estimator}) is a plug-in version of the objective function in (\ref{infeas estimator}) where we replace the Bellman error by $\gn(\cdot, \cdot; \eta, Q)$.  Compared to the classic empirical risk minimization, $(\hat \eta_n^\pi, \hat Q^\pi_n)$ solves a nested optimization problem in the sense that the objective function (\ref{estimator}) depends on $\gn(\cdot, \cdot; \eta, Q)$ which itself is the solution of another, lower-level optimization (\ref{est Bellman error}).   

The penalty term, $\lambda_n J_1^2(Q)$, is used to balance between the model fitting (i.e., the squared estimated Bellman error) and the complexity of the  relative value function measured by $J_1(Q)$. Similarly, $\mu_n J^2_2(g)$ is used to control the overfitting in estimating the {projected} Bellman error when the function class, $\G$, is complex. In the case where the function space is $k$-th order Sobolev space, the regularizer is typically defined by the $k$-th order derivative to capture the smoothness of function. In the case where the function space is Reproducing Kernel Hilbert Space (RKHS), the regularizer is the endowed norm.  In Supplement D, we provide a closed-form solution of the estimator when both $\Q$ and $\G$ are RKHSs. 

So far we have focused on evaluating a single target policy. In practice, one might want to compare the target policy to some reference policy or contrast multiple target policies of interest. Suppose we are interested in $K$ different target policies, $\left\{\pi_j\right\}_{j=1}^K$.  The  above procedure (\ref{estimator}) can be applied to estimate $\left\{\eta^{\pi_j}\right\}_{j=1}^K$. In the next section, we will provide the result of the joint asymptotic distribution of $\left\{\hat \eta_n^{\pi_j}\right\}_{j=1}^K$ (see Corollary \ref{theorem: multiple policies}). This can be used, for example, to construct the confidence interval of the difference of the average rewards between two policies.

\section{Theoretical Results}
\label{section: theory}

In this section, we first derive the global rate of convergence for  $(\hat \eta_n^\pi, \hat Q_n^\pi)$ in (\ref{estimator}) and derive the asymptotic distribution of $\hat \eta_n^\pi$ for a single policy. We then extend the results to the case of multiple policies.  For any state-action function, $f(s, a, s')$, and distribution, $\nu$, on $\S \times \A$, denote the $L_2(\nu)$ norm by $\norm{f}_{\nu}^2 = \int f^2(s, a) d\nu(s, a)$. If the norm does not have a subscript, then the expectation is with respect to the average state-action distribution in the trajectory, $\D$, that is,  $\norm{f}^2 = \EE\left\{(1/T) \sum_{t=1}^{T}f^2(S_t, A_t)\right\}$. 

We first state two standard assumptions used in the non-parametric regression literature \citep{gyorfi2006distribution}. Recall that the shifted relative value function is defined as $\tilde Q^\pi = Q^\pi - Q^\pi(s^*, a^*)$.

\begin{assumption}
	\label{assumption: bounded reward and value}
	The reward is uniformly bounded: $\abs{R_{t+1}} \leq R_{\max} < \infty$ for all $t \geq 1$. The shifted relative value function is bounded: $\abs{\tilde Q^\pi(s, a)} \leq Q_{\max}$ for all $s \in \S$ and $a \in \A$. 
\end{assumption}

\begin{assumption}
	\label{assumption: V requirement}
	The function class, $\Q$, satisfies (i) $Q(s^*, a^*) = 0$ and $\norm{Q}_{\infty} \leq Q_{\max}$ for all $Q \in  \Q$ and (ii) $\tilde Q^\pi \in \Q$.	
\end{assumption}

The assumption of a bounded reward is mainly to simplify the proof and can be relaxed to the sub-Gaussian case, that is, the error $R_{t+1} - r(S_t, A_t)$ is sub-Gaussian for all $t\le T$. The boundedness assumption on the shifted relative value function can be ensured by assuming certain smoothness assumptions on the transition distribution \citep{ortner2012online} or assuming geometric convergence to the stationary distribution \citep{hernandez2012further}.  The boundedness assumption,~\textit{(i)}, for members of  the function class, $\Q$, is used to simplify the proof; a truncation argument can be used to avoid this assumption.  

Recall that  $\g(\cdot, \cdot; \eta, Q)$ is a projected Bellman error in (\ref{surrogate BelErr Op}) into a function class, $\G$.  We make the following assumptions about $\G$.  

\begin{assumption}
	\label{assumption: G requirement}
	The function class, $\G$, satisfies 
	(i) $0 \in \G$, 
	(ii) $\norm{g}_{\infty} \leq G_{\max}$ for all $g \in \G$, and 
	(iii) $\kappa = \inf \bigdkh{\norm{\g (\cdot, \cdot; \eta, Q)}: \norm{\T_\pi(\cdot, \cdot; Q)-\eta- Q(\cdot, \cdot)} = 1, \eta \in \R, Q \in  \Q } > 0$.		
\end{assumption}

Given $R_{\max}$ and $Q_{\max}$, $G_{\max}$ can be chosen as $2(R_{\max}+Q_{\max})$.  Similar to $\Q$, the boundedness assumption of $\G$ is used to simply the proof and can be relaxed. The value of $\kappa$ measures how well the function class, $\G$, approximates the Bellman error for all $(\eta, Q)$ in which $\eta \in \R$ and $Q \in  \Q$.  The condition of a strictly positive $\kappa$ ensures the estimator (\ref{estimator}) based on minimizing the projected Bellman error onto the space, $\G$, is able to identify the true values, $(\eta^\pi, \tilde Q^\pi)$.  This is similar to the eigenvalue condition (Assumption 5) in \cite{luckett2019estimating}, but they are essentially using the same function class for $\Q$ and $\G$. 
Recall that, unlike in \cite{farahmand2016regularized}, here we do not assume $\T_\pi (\cdot, \cdot; Q)-\eta- Q(\cdot, \cdot)\in \G$  for every $(\eta, Q)$. If this were the  case, then we would have  $\g (\cdot, \cdot; \eta, Q)=\T_\pi(\cdot, \cdot; Q)-\eta- Q(\cdot, \cdot)$ and thus $\kappa = 1$.  

Below we make assumptions on the complexity of the function classes, $\Q$ and $\G$. These assumptions are satisfied  for common function classes, for example RKHS and Sobolev spaces \citep{van2000empirical,zhao2016partially,steinwart2008support,gyorfi2006distribution}.  We  denote  by $\cal N(\epsilon, \F, \norm{\cdot})$ the $\epsilon$-covering number of a set of functions, $\F$, with respect to the norm, $\norm{\cdot}$. 

\begin{assumption}
	\label{assumption: bounding J(EQ)}
	\label{assumption: entropy}
	(i) The regularization functional $J_1$ and $J_2$ are pseudo norms and induced by the inner products $J_1(\cdot, \cdot)$ and $J_2(\cdot, \cdot)$, respectively. There exist constants $C_1, C_2$ such that $J_2(\g(\cdot, \cdot; \eta, Q)) \leq C_1 + C_2 J_1(Q)$ holds for all $(\eta, Q) \in \R \times Q$. 
	(ii) Let $\Q_M = \{c + Q: \abs{c} \leq R_{\max}, Q \in \Q, J_1(Q) \leq M\}$ and $\G_M = \{g: g \in \G, J_2(g) \leq M\}$. There exist constants $C_3$ and $\alpha \in (0, 1)$, such that for any $\epsilon, M > 0$, 
	\begin{align*}
	& \max \big\{\log {\cal N}_{} (\epsilon, \G_M, \norm{\cdot}_\infty), \log {\cal N}(\epsilon, \Q_M, \norm{\cdot}_\infty)\big\} \leq C_3\left(\frac{M}{\epsilon} \right)^{2\alpha}.
	\end{align*}
\end{assumption}
The upper bound on $J_2(\g(\cdot, \cdot; \eta, Q))$ in \textit{(i)} is realistic when the transition kernel is sufficiently smooth (see \cite{farahmand2016regularized} for an example of MDP satisfying this condition).  We use a common $\alpha \in (0, 1)$ for both $\Q$  and $\G$ in \textit{(ii)} to simply the proof.

Now we are ready to state the theorem about the convergence rate for $(\hat \eta_n^\pi, \hat Q_n^\pi)$ in terms of the Bellman error. 

\begin{theorem}[Global Convergence Rate] \label{theorem: global rate}
	
	Let $(\hat \eta_n^\pi, \hat Q_n^\pi)$ be the estimator defined in (\ref{estimator}). Suppose Assumptions 1-5 hold and the tuning parameters, $(\lambda_n, \mu_n)$, satisfy $\tau^{-1}n^{-\frac{1}{1+\alpha}} \leq \mu_n \leq \tau \lambda_n$ for some constant, $\tau>0$.  Then the following bounds hold with probability at least $1-\delta$,  
	\begin{align*}
	& \bignorm{\T_\pi(\cdot, \cdot; \hat Q_n^\pi)-\hat \eta_n^\pi- \hat Q_n^\pi(\cdot, \cdot )}^2 \leqconst \kappa^{-2} \lambda_n (1+J_1^2(\tilde Q^\pi)) (1+\log(1/\delta)), \\
	& J_1(\hat Q_n^\pi) \leqconst 1+\log(1/\delta)  + J_1^2(\tilde Q^\pi),
	\end{align*}
	where the leading constants depend only on $(\tau, R_{\max}, Q_{\max}, G_{\max},  C_1, C_2, C_3, \alpha)$. 	
\end{theorem}

In Lemma  \ref*{lemma: upper bound of eta, Q} in Supplement B, we show that up to a constant, $\abs{\hat \eta_n^\pi - \eta^\pi} \leqconst \norm{\T_\pi(\cdot, \cdot ;\hat Q_n^\pi)-\hat \eta_n^\pi- \hat Q_n^\pi(\cdot, \cdot )}^2
$ and thus $\hat \eta_n^\pi$ is a consistent estimator for $\eta^\pi$ when $\lambda_n = o_P(1)$.  
When the tuning parameters are chosen such that $\lambda_n \asymp \mu_n$ and $\lambda_n  \asymp n^{-1/(1+\alpha)}$, the Bellman error at $(\hat \eta_n^\pi, \hat Q_n^\pi)$ has the optimal rate of convergence, i.e., $\norm{\T_\pi(\cdot, \cdot ;\hat \eta_n^\pi, \hat Q_n^\pi)-\hat \eta_n^\pi- \hat Q_n^\pi(\cdot, \cdot )}^2= O_P(n^{-1/(1+\alpha)})$. The proof of Theorem \ref{theorem: global rate} is provided in Supplement B.

In the following, we provide the asymptotic distribution of the estimated average reward.   This requires additional notation as follows.
Define $d^\pi(s, a) = \pi(a\given s) d^\pi(s)$;  $d^\pi$ is the density of the stationary distribution of the state-action under the target policy, $\pi$. For each $t \geq 1$,  denote by $d_t(s, a)$ the density of the state-action pair in the trajectory, $\D$, under the behavior policy.  Let $\bar d_T(s, a)$ be the average density over $T$ decision times.  
Motivated by the least favorable direction in partial linear regression problems \citep{van2000empirical,zhao2016partially},  we define the direction function, $e^\pi(s, a)$, by
\begin{align}
e^\pi(s, a) = \frac{d^\pi( s,  a)/\bar d_T( s,  a)}{\int (d^\pi( \tilde s, \tilde a)/\bar d_T(\tilde s, \tilde a)) d^\pi(\tilde s, \tilde a) d\tilde sd \tilde a }. \label{epi}
\end{align}
The  direction, $e^\pi$, is used to control the bias $(\hat \eta_n^\pi - \eta^\pi)$ caused by the penalization on the non-parametric component (i.e., relative value function) in the estimator (\ref{estimator}). This is akin to partially linear regression problem, $Y = f(Z) + X^\transpose \beta + \epsilon$,  in which the analog of $e^\pi(s, a)$ is the residual $x - \EE(X\given Z=z)$ (see \cite{donald1994series,van2000empirical} for the analysis in the regression problem).  
In our setting, the direction, $e^\pi(s, a)$, satisfies the following orthogonality: for any state-action function, $Q$,
\begin{align}
\EE \left[\frac{1}{T}\sum_{t=1}^T e^\pi(S_t, A_t) \left\{Q(S_t, A_t) - \sum_{a'} \pi(a'\given S_{t+1}) Q(S_{t+1}, a')\right\}  \right]= 0. \label{orthogonality}
\end{align}
To see this, note that $\int Q(s,a) d^\pi(s, a) dsda = \int \sum_{a'} \pi(a'\given s') Q(s',a') P(s'\given s, a) d^\pi(s, a) dsdads'$. 
The numerator in (\ref{epi}) is a ratio between the stationary distribution of state-action pair under target policy, $\pi$, and the average distribution of state-action pair in the trajectory, $\D$, under the behavior policy. The denominator is  the expectation of the ratio under the stationary distribution.   As a result of the denominator, we have $\int e^\pi(s, a) d^\pi(s, a) dsda = 1$.

Next define $q^\pi(s, a) = 	\operatorname*{lim}_{t^* \goes \infty} \frac{1}{t^*}\sum_{t=1}^{t^*} \EE_\pi\left[\sum_{k=1}^{t} \left\{1 - e^\pi(S_{k}, A_{k})\right\} \given S_1 = s, A_1 = a\right]$.  Note $q^\pi$ has a similar structure to that of the relative value function (\ref{value function. simplified}) in which the ``reward'' at time, $t$, is $\left\{1-e^\pi(S_t, A_t)\right\}$ and the ``average reward'' is zero (i.e., $\int \{1 - e^\pi(s, a)\} d^\pi(s, a) dsda = 0$).  Similar to the relative value function (\ref{value function. simplified}), $q^\pi(\cdot, \cdot)$ satisfies a Bellman-like equation:
\begin{align}
q(s, a) = 1- e^\pi(s, a) + \EE \left\{\sum_{a'} \pi(S_{t+1}, a') q(S_{t+1}, a') \Given S_t = s, A_t = a\right\}.\label{epi and qpi}
\end{align}

We make the following smoothness assumption about $e^\pi$ and $q^\pi$, akin to the assumptions used in partially linear regression literature \citep{van2000empirical,zhao2016partially}.

\begin{assumption}
	\label{assumption: qpi}
	The shifted function, $\tilde q^\pi = q^\pi - q^\pi(s^*, a^*) \in \Q$ and $e^\pi \in \G$.  
\end{assumption}

Recall that in Assumption \ref{assumption: V requirement} we restrict $Q(s^*, a^*) = 0$ for all $Q \in \Q$. Thus we consider the shifted function $\tilde q^\pi$ in the assumption above.  The analog of $\tilde q^\pi \in \Q$ in  partially linear regression problem, $Y =f(Z) + X^\transpose \beta  + \epsilon$, is the standard assumption that  $\EE[X|Z=\cdot] \in \F$, where $\F$ is the function class to model the nonparametric component, $f(z)$ \citep{donald1994series,van2000empirical}. The condition, $\tilde q^\pi \in \Q$, will be used to prove the $\sqrt{n}$ rate of convergence and asymptotic normality of $\hat \eta_n^\pi$. 
On the other hand, unlike in the regression setting, we assume that the  direction function, $e^\pi$, is sufficiently smooth (i.e., $e^\pi \in \G$).  
This assumption will be used to show that the bias of the coupled estimator, $\hat \eta^\pi_n$, decreases sufficiently fast to zero.

The last assumption is a contraction-type property. This assumption will be used to control the variance of a remainder term caused by the estimation of $Q^\pi$.   
\begin{assumption}
	\label{assumption: contraction}
	Let $(\P^\pi f)(\cdot, \cdot) = \EE_\pi\left\{f(S_{t+1}, A_{t+1})\given S_t=\cdot, A_t = \cdot\right\}$ be the function of the conditional expectation and $\mu^\pi(f) = \int f(s, a) d^\pi(s, a) ds da$ be the expectation under stationary distribution induced by $\pi$ for a state-action function, $f$. There exist constants, $C_4 > 0$ and $0 \leq \beta < 1$, such that for $f \in L_2$ and $t \geq 1$,
	\begin{align}
	\norm{(\P^\pi)^t (f) - \mu^\pi(f)}  \leq C_4 \norm{f}_{} \beta^t. \label{ass}
	\end{align}
	
\end{assumption}
The parameter, $\beta$, in Assumption \ref{assumption: contraction} is akin to the discount factor, $\gamma$, in the discounted reward setting. Intuitively, this is related to the ``mixing rate'' of the Markov chain induced by the target policy $\pi$. A similar assumption was imposed in \cite{van1998learning} (Assumption 7.2 on p.~99). Now we are ready to present our main result, the asymptotic normality of the estimated average reward, $\hat \eta_n^\pi$.

\begin{theorem}[Asymptotic Distribution] \label{theorem: normal} 
	
	Suppose the conditions in Theorem \ref{theorem: global rate} hold. In addition, suppose Assumption \ref{assumption: qpi} and \ref{assumption: contraction} hold and  $\lambda_n = a_n n^{-1/2}$ with $a_n \goes 0$.  The estimator, $\hat \eta_n^\pi$, in (\ref{est Bellman error}) is $\sqrt{n}$-consistent and asymptotically normal: $\sqrt{n} (\hat \eta_n^\pi - \eta^\pi) \wcvg 
	\textbf{N}(0, \sigma^2)$, where 
	\begin{align*}
	\sigma^2 = \operatorname{Var}\left[\frac{1}{T} \sum_{t=1}^{T}  \frac{d^\pi(S_t, A_t)}{\bar d_T(S_t, A_t)}\left\{R_{t+1} + \sum_{a'} \pi(a' \given S_{t+1})Q^\pi(S_{t+1}, a') - \eta^\pi - Q^\pi(S_t, A_t)\right\} \right]. 
	\end{align*}
\end{theorem}

From Theorem \ref{theorem: normal},  the variance in estimating the average reward parameter, $\eta^\pi$, depends on the length of trajectory and the ratio between the stationary distribution of the state-action pair induced by the target policy (i.e., $d^\pi$) and the average state-action distribution in the training data (i.e., $\bar d_T$).  To gain  intuition of how these impact the asymptotic variance of $\hat \eta^\pi_n$, consider a simplified setting where the conditional variance of $R_{t+1} + \sum_{a'} Q^\pi(S_{t+1}, a') - \eta^\pi - Q^\pi(S_t, A_t)$ given $(S_t, A_t)$ is a constant, denoted by $\sigma_{0}^2$.  
It can be shown that the asymptotic variance becomes 
$
\sigma^2 = \frac{\sigma_{0}^2}{T}(1+ \norm{(d^\pi/\bar d_T)-1}^2)
$.
Thus the smaller $\norm{(d^\pi/\bar d_T)-1}^2$  (i.e., the ratio, $d^\pi/\bar d_T$, close to one),  the smaller  the asymptotic variance of the estimated average reward.    Although here we focus only on the asymptotic properties of $\hat \eta_n^\pi$ for large $n$ (recall $n$ is the number of i.i.d.~trajectories), one can see that increasing length of the trajectory, $T$, reduces the asymptotic variance.

Now we present the result for evaluating a class of policies, $\Pi = \{\pi_1, \dots, \pi_K\}$. Denote by $\hat \eta_n^{\pi_j}$ the estimated average reward of the policy, $\pi_j$, using (\ref{estimator}).  
\begin{corollary}[Multiple Policies] \label{theorem: multiple policies}
	Suppose the conditions in Theorem \ref{theorem: global rate} and \ref{theorem: normal} hold for each $\pi \in \Pi$. Let $\epsilon_t^{\pi} = \frac{d^{\pi}(S_t, A_t)}{\bar d_T(S_t, A_t)}[R_{t+1} + \sum_{a'} \pi(a'\given S_{t+1}) Q^{\pi}(S_{t+1}, a') - \eta^{\pi} - Q^{\pi}(S_t, A_t)]$ for each $\pi \in \Pi$. Then the estimated average rewards, $\{\hat \eta_n^{\pi_1}, \dots, \hat \eta_n^{\pi_K}\}$,  jointly converge in distribution to a multivariate Gaussian distribution:
	\begin{align*}
	\begin{pmatrix}
	\sqrt{n}(\hat \eta_n^{\pi_1} - \eta^{\pi_1}) \\
	\vdots\\
	\sqrt{n}(\hat \eta_n^{\pi_K}  - \eta^{\pi_K})
	\end{pmatrix} \wcvg \textbf{MVN}(\boldmath{0}, \Sigma),
	\end{align*}
	where the $(i, j)$ element of $\Sigma$ is given by $\EE\left[\left\{(1/T) \sum_{t=1}^{T} \epsilon_t^{\pi_i} \right\}\left\{(1/T) \sum_{t=1}^{T}  \epsilon_{t}^{\pi_j} \right\}\right]$.
\end{corollary}

To conduct inference, we need to estimate the asymptotic variance, $\Sigma$. For each $\pi \in \Pi$, we denote the plug-in estimation of $\epsilon_t^\pi$ (defined in Corollary \ref{theorem: multiple policies}) by  $\hat \epsilon_t^\pi$ in which we plug in $(\hat \eta_n^\pi, \hat Q_n^\pi)$ and an estimator for the ratio, $d^\pi(s, a)/\bar d_T(s, a)$. We then estimate the asymptotic variance, $\Sigma$, by 
$\hat \Sigma_n = \left[\Pn \left\{\left(\frac{1}{T} \sum_{t=1}^{T} \hat \epsilon_t^{\pi_i}\right) \left(\frac{1}{T} \sum_{t=1}^{T} \hat \epsilon_t^{\pi_j}\right)\right\}\right]_{i, j=1}^K$. 

We can estimate the ratio, $d^\pi/\bar d_T$, as follows.  First, we note that by taking the expectation on both sides of (\ref{epi}), the ratio can be written in terms of $e^\pi$: $d^\pi(s, a)/\bar d_T(s, a) = e^\pi(s, a) /\EE\{(1/T)\sum_{t=1}^{T} e^\pi(S_t, A_t)\}.$
It is enough to construct an estimator for $e^\pi$, which we denote by $\hat e_n^\pi$, and then estimate the ratio by $\hat e_n^\pi(s, a) /\Pn\{(1/T)\sum_{t=1}^{T} \hat e_n^\pi(S_t,A_t)\}$. 
Motivated by the orthogonality (\ref{orthogonality}) and the expression (\ref{epi and qpi}),  we construct the estimator for $e^\pi(\cdot, \cdot)$ by $\hat e^\pi_n(\cdot, \cdot) = \tilde g_{n, \pi}(\cdot,\cdot;\hat q_n^\pi)$, where 
$\hat q_n^\pi(\cdot,\cdot) = \argminb_{q \in \Q} \Pn\{(1/T)\sum_{t=1}^{T}  \tilde g_{n, \pi}^2(S_t, A_t; q)\} + \tilde \lambda_n J_1^2(q)$ and $\tilde g_{n, \pi}(\cdot, \cdot; q) = \argminb_{g \in \G} \Pn[(1/T)\sum_{t=1}^{T} \{1-q(S_{t}, A_t) + \sum_{a'} \pi(a'\given S_{t+1}) q(S_{t+1}, a')- g(S_t, A_t)\}^2]+ \tilde \mu_n J_2^2(g)$ for each $q \in \Q$. Here $(\tilde \lambda_n, \tilde \mu_n)$ are some tuning parameters.  Following a similar argument as in the proof of Theorem \ref{theorem: global rate}, $\tilde q^\pi_n$ can be shown to be a consistent estimator for $\tilde q^\pi$. Under the assumption that $e^\pi \in \G$, $e^\pi$ can be consistently estimated by $\hat e^\pi_n(\cdot, \cdot) = \tilde g_{n, \pi}(\cdot, \cdot; \hat q_n^\pi)$ based on (\ref{epi and qpi}).  See Supplement C for additional details about the estimator $\hat e^\pi_n$.   In Supplement D, we provide a closed-form solution for the estimator for  the asymptotic variance when $\Q$ and $\G$ are RKHSs.

\section{Simulation}
\label{section: simulation}

In this section, we conduct a simulation study to evaluate the performance of the proposed method.  The generative model is given as follows.  We follow the state generative model in \cite{luckett2019estimating}. Specifically, the state, $S_t = (S_{t,1}, S_{t, 2})$, is a two-dimensional vector and the action, $A_t \in \{0, 1\}$, is binary.  Given the current state, $S_t$, and action, $A_{t}$, the next state, $S_{t+1} = (S_{t+1, 1}, S_{t+1, 2})$, is generated by $S_{t+1, 1} =  (3/4)(2A_t -1)S_{t, 1} + (1/4) S_{t,1} S_{t, 2} + \boldsymbol{N}(0, 0.5^2)$ and $S_{t+1, 2} = (3/4)(1 - 2A_t)S_{t, 2} + (1/4) S_{t,1} S_{t, 2} +  \boldsymbol{N}(0, 0.5^2)$. Note that receiving a treatment $(A_t = 1)$ increases the value of $S_{t, 1}$ while decreases $S_{t, 2}$. The reward is generated by  $R_{t+1} =S_{t+1, 1} + (1/2) S_{t+1, 2} + (1/4) (2A_t - 1)$.   For each trajectory in the training data,  the state variables are generated as independent standard  normal random variables and the behavior policy is to choose $A_t = 1$ with a fixed probability 0.5.   We evaluate and compare two natural policies: the ``always treat'' policy, $\pi_1 (a\given s) = 1$, and ``no treatment'' policy, $\pi_2(a\given s) = 0$.  

In the implementation, we use RKHS with the radial basis function (RBF) kernel to construct the function classes, $\Q$ and $\G$. The details of how to modify an arbitrary RKHS such that the value at $(s^*, a^*)$ is zero can be found in Supplement D.  The bandwidth parameter in the RBF kernel is chosen by the median heuristic. Recall that the estimator (\ref{estimator}) involves two tuning parameters, $(\lambda_n, \mu_n)$.  Following the idea in \cite{farahmand2011model}, we select these tuning parameters as follows.  We first split the dataset into a training set, $\D^{\text{trn}}$,  and a validation set, $\D^{\text{val}}$. For each candidate value of the tuning parameters, $(\lambda, \mu)$, the training set, $\D^{\text{trn}}$, is used to form the estimator by  (\ref{estimator}) and (\ref{est Bellman error}).~Denote the corresponding estimator by $\left\{\hat \eta^\pi(\lambda, \mu), \hat Q^\pi(\cdot, \cdot; \lambda, \mu)\right\}$. Then the temporal difference (TD) error, $R + \sum_{a'} \hat Q^\pi(S', a'; \lambda, \mu) - \hat \eta^\pi(\lambda, \mu) - \hat Q^\pi(S, A; \lambda, \mu),$ is calculated for each transition sample, $(S, A, S', R)$, in the validation set, $\D^{\text{val}}$.   Recall that the Bellman error is zero at $(\eta^\pi,  \tilde Q^\pi)$.  We use the validation set, $\D^{\text{val}}$, to fit a model for the Bellman error with respect to $(\hat \eta^\pi(\lambda, \mu), \hat Q^\pi(\cdot, \cdot; \lambda, \mu))$ and denote the estimated Bellman error by $\hat f(\cdot, \cdot; \lambda, \mu)$. Note that this step is essentially a regression problem (i.e., the dependent variable is the TD error and independent variables are the current state and action).  Finally, we choose $(\lambda, \mu)$ that minimizes the squared estimated Bellman error over the validation set,  i.e., $\sum_{(S, A) \in \D^{\text{val}}} \hat f^2(S, A; \lambda, \mu)$.  The final estimator for $\eta^\pi$ is then calculated with the optimal tuning parameters using the entire dataset.  In the simulation, we use (1/2) of the trajectories for the training set and (1/2) for the validation set and we use Gaussian Process regression to estimate the Bellman error in the validation step.

We consider different scenarios of the number of the trajectories, $n \in \{25, 40\}$, and  the length of each trajectory, $T \in \{25, 50, 75\}$.  In each scenario, we generate 500 simulated dataset and for each dataset we construct the $95\%$ confidence intervals  of $\eta^{\pi_1}, \eta^{\pi_2}$ and $\eta^{\pi_1} - \eta^{\pi_2}$.  The coverage probability of each confidence interval is calculated over 500 repetitions.  The simulation result is reported in Table \ref{coverage}.   When the number of trajectories is small (i.e., $n = 25$), the simulated coverage probability is slightly smaller than the claimed value, 0.95, especially when the length of the trajectory, $T$, is small. It can be seen that the coverage probability slightly improves when $T$ increases. When $n = 40$, the coverage probability becomes closer to 0.95 as desired. Overall, the simulation result demonstrates the validity of the inference and the selection procedure  for the tuning parameters. It suggests that it is necessary to perform a small-sample correction when both $n$ and $T$ are small. This is left for future work.

\begin{table}
	\caption{Coverage probability of the $95\%$ confidence interval  and MAD (mean absolute deviation) over 500 repetitions. Case 1: policy evaluation of $\pi_1$. Case 2: policy evaluation of $\pi_2$. Case 3: policy comparison between $\pi_1$ and $\pi_2$. }	
	\vspace{2ex}
	\centering
	\begin{tabular}{c|c|c|c|c||c|c|c|c}
		\hline
		& $n$  & $T$  & Coverage Prob. & MAD    & $n$  & $T$  & Coverage Prob. & MAD    \\ \hline
		\multirow{3}{*}{Case 1} & 25 & 25 & 0.926 & 0.0702 & 40 & 25 & 0.944 & 0.0546 \\ 
		& 25 & 50 & 0.930 & 0.0535 & 40 & 50 & 0.944 & 0.0427 \\ 
		& 25 & 75 & 0.938 & 0.0438 & 40 & 75 & 0.948 & 0.0346 \\ 
		\hline
		\multirow{3}{*}{Case 2} & 25 & 25 & 0.934 & 0.0368 & 40 & 25 & 0.928 & 0.0313 \\ 
		& 25 & 50 & 0.946 & 0.0261 & 40 & 50 & 0.940 & 0.0224 \\ 
		& 25 & 75 & 0.922 & 0.0222 & 40 & 75 & 0.942 & 0.0185 \\ 
		\hline
		\multirow{3}{*}{Case 3} & 25 & 25 & 0.932 & 0.0761 & 40 & 25 & 0.946 & 0.0612 \\ 
		& 25 & 50 & 0.928 & 0.0598 & 40 & 50 & 0.948 & 0.0461 \\ 
		& 25 & 75 & 0.932 & 0.0480 & 40 & 75 & 0.948 & 0.0388 \\ 
		\hline
	\end{tabular}
	
	\label{coverage}
\end{table}

\section{Case Study: HeartSteps}
\label{section: heartsptes}

We apply the method to the data collected in the first study in HeartSteps \citep{klasnja2015microrandomized,Liaoetal2015,klasnja2018efficacy}. Below we refer to this study by HS1 for simplicity.  HS1 was a 42-day MRT with 44 healthy sedentary adults.  We focus on the activity suggestion intervention component. There were five individual-specified times in a day which were roughly separated by 2.5 hours and corresponded to the individual’s morning commute, mid-day, mid-afternoon, evening commute, and post-dinner times. At each decision time, an activity suggestion was sent with a fixed probability 0.6 only if the participants were considered to be available for treatment.  For example, the participants were considered unavailable when they were currently physically active (e.g., walking or running) or driving a vehicle.  The activity suggestions were intended to motivate near-time walking. Each participant wore a Jawbone wrist tracker and the minute-level step count data was recorded.

We construct the state based on the participant's step count data (e.g., the 30-min step count prior to the decision time and the total step count from yesterday), location, temperature and number of the notifications received over the last seven days. We also include in the state the time slot index in the day (1 to 5) and the indicator measuring how the step count varies at the current time slot over the last seven days.  The reward is formed by the log transformation of the total step count collected in 30-min window after the decision time.  The log transformation is performed as the step count data is positively skewed \citep{klasnja2018efficacy}.  The step count data might be missing because the Jawbone tracker recorded data only when there were steps occurred. We use the same imputation procedure as in \cite{klasnja2018efficacy}.  The state related to the step count are constructed based on the imputed step counts. The variables in the state are chosen to be predictive of the reward. In particular, each variable is selected, at the significance level of 0.05, based on a marginal Generalized Estimating Equation (GEE) analysis.  In the analysis, we exclude seven participants' data as in the primary analysis in \cite{klasnja2018efficacy} (three due to technical issues and four due to early dropout). In addition, from the 37 participants' data we exclude the decision times when participants were traveling abroad or experiencing technical issues or  when the reward (i.e., post 30-min step count) is considered as missing (see \cite{klasnja2018efficacy} for details).

We consider three target policies. The first policy, $\pi_{\text{nothing}}$, is ``do nothing''.  The second policy, $\pi_{\text{always}}$, is the ``always treat'' policy. Recall that in HeartSteps the activity suggestion can be sent only when the participant is available. So here the ``always treat'' policy refers to always send the suggestion whenever the participant is available.   The third policy, $\pi_{\text{location}}$, is based on the location. Specifically, we consider the policy that sends the activity suggestion when the participant is at either home or work location and available. This policy is of interest because people  at home or work are in a more structured environment and thus might be  able to better respond to an activity suggestion as compared with at other locations.  In HS1, about $44\%$~of the  available decision times were at times that  the participants were at their home or work location.  Thus the policy, $\pi_{\text{location}}$, is different from the ``always treat'' policy, $\pi_{\text{always}}$.

In the implementation, we use the RKHS with the radial basis function kernel to form the function classes, $\Q$ and $\G$.  The tuning parameters are selected based on the procedure described in Section \ref{section: simulation}.  
The estimated average reward of the location-based policy, $\pi_{\text{location}}$, is $3.155$ with the 95\% confidence interval , $[2.893, 3.417]$, which is  slightly better than the ``do nothing'' policy. Specifically, the estimated average reward of $\pi_{\text{nothing}}$ is $2.962$ and the 95\% confidence interval of the difference, $\eta^{\pi_{\text{location}}} - \eta^{\pi_{\text{nothing}}}$, is $[-0.016,  0.402]$.  Translating back to the raw step count as in \cite{klasnja2018efficacy}, the location-based policy is able to increase the average 30-min step count roughly by $22\%$ (i.e., $\exp(3.16 - 2.96) -1 = 1.22$), corresponding to $55$ steps (the mean post-decision time step count is 248 across all decision times in the dataset).  However if we compare the ``always treat'' policy ($\hat \eta^{\pi_{\text{always}}} = 3.127$, 95\%~confidence interval is $[2.840, 3.413]$) with the location-based policy, $\pi_{\text{location}}$, we see no indication that providing treatment only at home or work is better than always providing treatment (the 95\%~confidence interval of $\eta^{\pi_{\text{location}}}-\eta^{\pi_{\text{always}}} $ is $[-0.161,  0.217]$).   Recall that the sample size for this study is $n=37$ thus this non-significant finding may be due to the small sample.

\section{Discussion}
\label{sec: discussion}

In this work we developed a flexible method to conduct inference about the  the long-term average outcomes for given target policies using data collected from a possibly different behavior policy. We believe that this is an important first step towards developing data-based just-in-time adaptive interventions. Below we discuss some directions for future research.

In many MRT studies, a natural choice of the proximal outcome to assess the effectiveness of the intervention is binary.  For example, in the Substance Abuse Research Assistance study \citep{rabbi2018toward}, the proximal outcome was whether the individual completed a daily survey. An interesting open question is how to extend the method to the binary reward setting, which would require carefully choosing the model to represent the relative value function and/or the loss functions used in estimating the Bellman error and solving the Bellman equation.

Non-stationarity occurs mainly because of the unobserved aspects of the current state (e.g., the engagement and/or burden) in many mHealth applications. It will be interesting to generalize the average reward framework to incorporate the non-stationarity detected in the observed trajectory.  Alternatively, one can consider evaluating the treatment policy in the indefinite horizon setting where there is an absorbing state (akin to the individual disengaging from the mobile app) and thus we aim to conduct inference about the expected total rewards until the absorbing state is reached.  

We focused on evaluating and contrasting multiple pre-specified treatment policies. An important next step is to extend the method to learn the optimal policy that would lead to the largest long-term average reward and to develop the inferential methods to assess the usefulness of certain variables in the policy.

	\section*{Acknowledgment}
	
	This work was supported by National Institute on Alcohol Abuse and Alcoholism (NIAAA) of the National Institutes of Health under award number R01AA23187, National Institute on Drug Abuse (NIDA) of the National Institutes of Health under award numbers P50DA039838 and R01DA039901, National Institute of Biomedical Imaging and Bioengineering (NIBIB) of the National Institutes of Health under award number U54EB020404, National Cancer Institute (NCI) of the National Institutes of Health under award number U01CA229437, and National Heart, Lung, and Blood Institute (NHLBI) of the National Institutes of Health under award number R01HL125440. The content is solely the responsibility of the authors and does not necessarily represent the official views of the National Institutes of Health.

\bibliographystyle{agsm}
\bibliography{reference}
\endgroup

\end{document}